\documentclass[letterpaper, 10 pt, conference]{ieeeconf}  % Comment this line out if you need a4paper
\usepackage{graphicx}
\usepackage{makecell}
\usepackage{amssymb}% http://ctan.org/pkg/amssymb
\usepackage{pifont}% http://ctan.org/pkg/pifont
\usepackage{hyperref}

\newcommand{\cmark}{\ding{51}}%
\newcommand{\xmark}{\ding{55}}%

\IEEEoverridecommandlockouts                              % This command is only needed if
\title{\LARGE \bf
Forecasting Pedestrian Trajectory with Machine-Annotated Training Data
}

\author{Olly Styles$^{1}$, Arun Ross$^{2}$ and Victor Sanchez$^{1}$% <-this % stops a space
\thanks{$^{1}$Olly Styles and Victor Sanchez are with the Department of Computer Science,
        University of Warwick, Coventry, UK
        {\tt\small \{o.c.styles | v.f.sanchez-silva\}@warwick.ac.uk}}%
\thanks{$^{2}$Arun Ross is with the Department of Computer Science and Engineering, Michigan State University, East Lansing, USA
        {\tt\small  rossarun@cse.msu.edu}}%
}

\begin{document}

\maketitle
\thispagestyle{empty}
\pagestyle{empty}

%%%%%%%%%%%%%%%%%%%%%%%%%%%%%%%%%%%%%%%%%%%%%%%%%%%%%%%%%%%%%%%%%%%%%%%%%%%%%%%%
\begin{abstract}

Reliable anticipation of pedestrian trajectory is imperative for the operation of autonomous vehicles and can significantly enhance the functionality of advanced driver assistance systems. While significant progress has been made in the field of pedestrian detection, forecasting pedestrian trajectories remains a challenging problem due to the unpredictable nature of pedestrians and the huge space of potentially useful features. In this work, we present a deep learning approach for pedestrian trajectory forecasting using a single vehicle-mounted camera. Deep learning models that have revolutionized other areas in computer vision have seen limited application to trajectory forecasting, in part due to the lack of richly annotated training data. We address the lack of training data by introducing a scalable machine annotation scheme that enables our model to be trained using a large dataset without human annotation. In addition, we propose Dynamic Trajectory Predictor (DTP), a model for forecasting pedestrian trajectory up to one second into the future. DTP is trained using both human and machine-annotated data, and anticipates dynamic motion that is not captured by linear models.  Experimental evaluation confirms the benefits of the proposed model.

\end{abstract}

%%%%%%%%%%%%%%%%%%%%%%%%%%%%%%%%%%%%%%%%%%%%%%%%%%%%%%%%%%%%%%%%%%%%%%%%%%%%%%%%
\section{INTRODUCTION}

Interacting with humans in complex urban environments remains a challenging problem for autonomous vehicles (AVs). Unlike highways with well-defined rules for traffic, urban environments necessitate that vehicles interact with other road users, such as pedestrians and cyclists, in a more nuanced manner. For an AV to navigate effectively in such environments, the vehicle must be able to locate and react to pedestrians in order to avoid collisions. The first component of such a navigation system, detecting pedestrians, has seen a tremendous amount of research effort in the past decade \cite{tenyearsofdetection}. If current trends continue, performance will soon match and even surpass human-level performance on standard evaluation benchmarks \cite{howfarfromsolving}. The rapid advancements in this area have led to real-world implementations of advanced driver assistance systems (ADAS) to aid drivers in critical situations. Such systems are capable of providing warnings or initiating braking if a pedestrian is detected in front of the vehicle, but are less reliable in the anticipation of potentially dangerous events before a pedestrian steps into the roadway.

As vehicles move towards ever greater autonomy, the need for accurate pedestrian trajectory forecasting also grows. With a human driver in the loop, ADAS may be designed conservatively as false negatives can be tolerated. For an AV, however, the reliable anticipation of pedestrian \textit{intent} is a critical safety feature but a complex challenge. Although driven by long-term motion goals such as reaching a specific destination \cite{goaldirected}, pedestrian motion is highly dynamic and may change at a moment's notice, such as a child running rapidly into the street. To deal with this uncertainty, human drivers use heuristics such as pedestrian head pose, gait, and scene dynamics to reason about intent \cite{agreeing}. Without these cues, for example, human drivers find it more challenging to predict if a pedestrian is about to cross the road \cite{atthekerb}.

\begin{figure}[t]
\begin{center}
\includegraphics[width=0.92\linewidth]{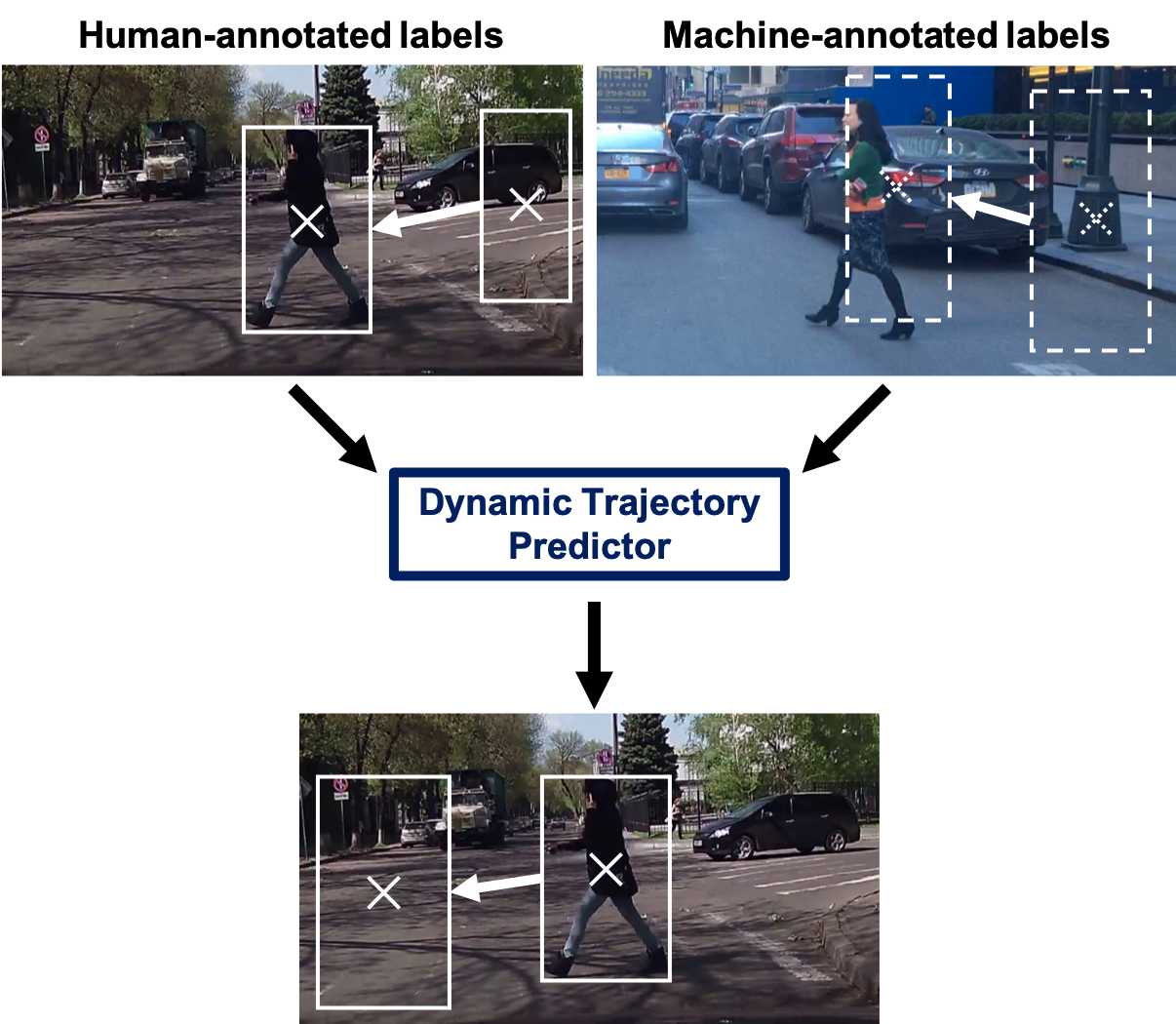}
\end{center}
       \vspace{-2mm}
   \caption{We propose a model and training regime for pedestrian trajectory forecasting. Due to a lack of annotated training data, our model is trained jointly with human-annotated and machine-annotated pedestrian bounding boxes generated by a pedestrian detection and tracking algorithm.}
\label{fig:pull}
\vspace{-2mm}
\end{figure}

Modern vehicles equipped with sensors such as LiDAR and Radar can build an accurate representation of the surrounding environment \cite{perception-survey}. Both LiDAR and Radar, however, lack the capability for extracting high-resolution features and are, thus, commonly supplemented with visible spectrum cameras.  Manual annotation of features such as pedestrian head pose and body language cues from camera data is challenging and time-consuming. Furthermore, pedestrian behavior varies across different cultures and driving environments \cite{culturechapter}. A model trained to anticipate pedestrian behavior in California, USA is unlikely to perform well on the streets of Mumbai, India. Without a practical method for learning from unlabelled data, it is likely that large quantities of data must be manually annotated for deployment in each environment.

Based on the above observations, we present a system for pedestrian trajectory forecasting capable of learning from unlabeled data. The two main contributions of this work are as follows:

\begin{enumerate}
\item Dynamic Trajectory Predictor (DTP), a pedestrian trajectory forecasting deep learning model based on motion features from optical flow.
\item A machine annotation scheme for training trajectory forecasting models in the absence of labeled data.
\end{enumerate}

\section{RELATED WORK}
Our proposed approach builds on the substantial progress made in pedestrian detection and human action recognition. However, in this section, we concentrate on literature more directly relevant to our contributions, that are focused on (a) pedestrian trajectory forecasting and (b) alternative supervision methods for training models in the absence of large-scale human annotated datasets. For pedestrian detection, see recent surveys such as \cite{tenyearsofdetection,towardshuman}. For action recognition, see recent surveys such as \cite{ar-survey-1,ar-survey-2}.

\subsection{Pedestrian Trajectory Forecasting}

\textbf{Dynamic Systems Approach.} Given the absence of large pedestrian trajectory datasets, previous works have modeled the dynamic motion of pedestrian's using linear dynamic systems (LDS) that combine the assumptions of constant velocity (CV) or constant acceleration (CA) with a filtering algorithm such as the Kalman filter \cite{pathpredictionwithbayesian}. To model non-linear, dynamic motion, a switching linear dynamic system (SLDS) uses a discrete Markov chain to select between multiple LDS at each timestep based on past observations. However, the SLDS is limited to \textit{reacting} to pedestrian motion rather than \textit{anticipating} a change in dynamics. To address this issue, existing works \cite{context-based,gavrilaijcv} focus on additional cues such as pedestrian head pose, motion state, and road scene context or use a non-linear filtering algorithm such as the unscented Kalman filter \cite{ukf}.

\textbf{Data-Driven Approach.} Data-driven approaches for trajectory forecasting have gained attention in recent years resulting from the success of deep learning models for related problems such as image classification, action recognition, and pedestrian detection. In particular, deep learning models have been applied to trajectory forecasting in a surveillance setting with a fixed overhead camera on datasets such as UCY \cite{ucy} and ETH \cite{eth}, or forecasting vehicle trajectories \cite{egocentric}. In \cite{behavior-cnn}, pedestrian trajectory is forecast by encoding pedestrian location as a sparse vector which is used directly as input to a convolutional neural network (CNN).  In \cite{sociallstm}, pedestrian trajectory is forecast from a static, overhead camera using a long short-term memory (LSTM) network. The authors introduce social pooling, which models the social interactions between multiple pedestrians.

Trajectory forecasting is considered from a first-person perspective in \cite{fpl}. The authors propose a model combining features from the pedestrian’s pose, estimated ego-motion, and past location information. Similarly, in \cite{longterm}, an LSTM is used to predict the future location of pedestrian bounding boxes by first estimating future ego-motion and then using these estimates with observed bounding boxes to forecast the location of future bounding boxes. All data-driven approaches, however, are limited by the lack of available training data.

\subsection{Alternative supervision}

Supervised learning has been a prominent learning paradigm requiring accurate annotation of datasets, which is commonly completed manually through painstaking human effort. Due to the massive quantities of data necessary to effectively train state-of-the-art models, several alternative means of supervision have been proposed. Pre-training neural network models on the large Imagenet dataset \cite{imagenet}, before fine-tuning on a target dataset, has become the de facto standard in settings where annotated data is limited. Alternative means of building large annotated datasets for pre-training such as mining social media websites \cite{instahashtags} have also been proposed. An alternative learning paradigm is self-supervision. In self-supervision, some subset of a dataset is withheld during the training process, and a model is trained to predict the withheld data. In this way, a model may exploit large-scale datasets without expensive annotation. For example, the authors in \cite{colorization} convert color images to greyscale and train a model to perform the inverse operation, and the authors in \cite{patch-prediction} predict the location of image patches in relation to another patch. In an intelligent vehicle setting, existing works have used data collected by one sensor (such as a camera) to predict the data collected by another sensor (such as an inertial measurement unit) \cite{predict-ego, end-to-end-driving}. Self-supervision avoids the expensive human annotation component of supervised learning and is, therefore, well-suited to address problems with limited annotated data. Our proposed machine annotation scheme enables us to leverage the power of self-supervision for pedestrian trajectory forecasting.

\section{PROPOSED METHOD} \label{sec:3}

\subsection{Problem formulation and baseline} \label{formulation}

Consider a pedestrian localized in a video with an associated set of bounding box coordinates for the current and past $m$ frames, such as in Fig \ref{fig:pull}. Our goal is to predict the centroid of future bounding boxes with coordinates $x_t$ and $y_t$ as to anticipate potentially dangerous events, such as a pedestrian stepping into the roadway. The horizontal and vertical components of velocity, $v^x_t$ and $v^y_t$ respectively, at time $t$ of a pedestrian relative to the vehicle in the 2D projection obtained by a camera can be estimated by taking the first order derivative of the past centroids:

\begin{equation}
v^x_t = \frac{x_t - x_{t-m}}{m} \quad , \quad v^y_t = \frac{y_t - y_{t-m}}{m}
\end{equation}

As a baseline, we consider that the pedestrian maintains their average velocity of the previous $m$ timesteps in the future $n$ timesteps:

\begin{equation}
\widetilde{x}_{t+n} = x_t + v^x_t \cdot n \quad , \quad \widetilde{y}_{t+n} = y_t + v^y_t \cdot n
\end{equation}

We denote a pedestrian's centoid location at time $t$ as $L_t$, comprised of coordinates $x_t$ and $y_t$. Similarly, we denote velocity at time $t$ as $v_t$, comprised of vertical and lateral velocities $v^x_t$ and $v^y_t$. The predicted location of the centroid at time $t+n$ following the constant velocity assumption is denoted as:

\begin{equation} \label{cv}
\widetilde{L}_{t+n} = L_t + v_t \cdot n
\end{equation}

We focus here on predicting the centroid in the 2D coordinate space obtained by a camera, rather than the 3D world coordinates required for full localization by an AV. In practical applications, 2D object detections may be associated with 3D world coordinates using a depth estimation method such as \cite{stereomatching}.

\subsection{Dynamic Trajectory Predictor} \label{sec:dynamic-motion}

\begin{figure}[t]
\begin{center}
\includegraphics[width=.75\linewidth]{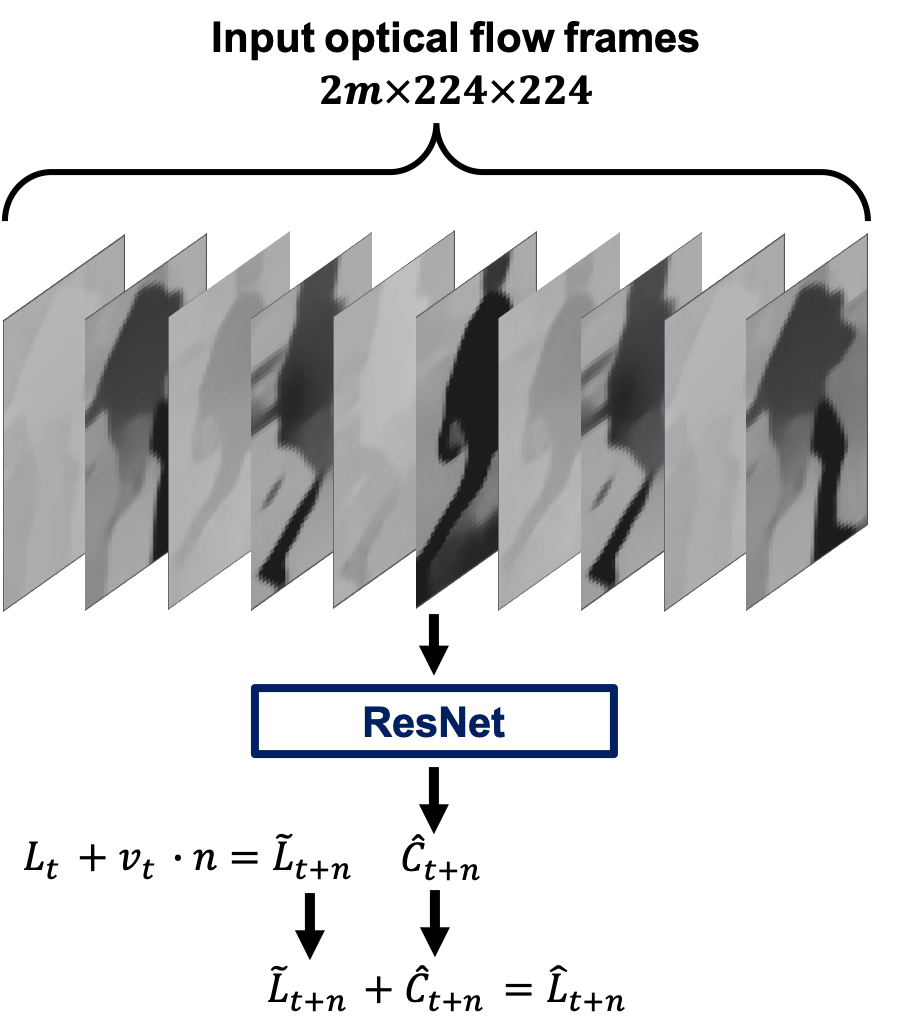}
\end{center}
       \vspace{-4.5mm}
   \caption{DTP forecasts pedestrian trajectory reletive a constant velocity baseline. We use ResNet \cite{resnet} with modified input and output layers to compute features from past optical flow. See Section \ref{sec:model} for details.}
\label{fig:model}
       \vspace{-3.5mm}
\end{figure}

In many scenarios, such as when a pedestrian is stationary or walking at a constant speed, the constant velocity assumption is a reasonable predictor of future location. Challenging situations are instances that deviate significantly from this assumption. An effective model must anticipate a change in velocity and adjusts predictions accordingly. The error resulting from the constant velocity assumption is denoted by:

\begin{equation}
\widetilde{e}_{t+n} = |L_{t+n} - \widetilde{L}_{t+n}|
\end{equation}

Rather than directly predicting a location $\hat{L}_{t+n}$ directly, existing works \cite{fpl, longterm} output the location relative to the last observed timestep, $\Delta L_{t+n} = L_{t+n} - L_{t}$. In contrast, we propose to output a compensation term, $C_t = - \widetilde{e}_t$, which corrects for errors in the constant velocity assumption. In this way, our model is first initialised to a strong baseline (in the case where $C_t = 0$, the model's predictions equal constant velocity) and then fine-tunes predictions on training examples for which the constant velocity assumption results in errors. The final predicted coordinates in the original 2D image projection, $\hat{L}_{t+n}$, are then recovered as follows:

\begin{equation}
 \hat{L}_{t+n}=\tilde{L}_{t+n} + \hat{C}_{t+n}
\end{equation}

Inspired by effective action recognition models \cite{twostream, tsn}, DTP uses a stack of optical flow frames as input to a CNN that extracts a compact representation of human motion. From this feature vector, a fully connected layer outputs a prediction $\hat{C}_{t+n}$ representing the estimated correction factor. A vector of large magnitude indicates that the pedestrian velocity will increase or decrease, whereas a vector of magnitude close to 0 indicates that the pedestrian will maintain their current velocity. We use ResNet \cite{resnet} as our backbone network, owing to its consistently good performance on many vision tasks. A high-level diagram of our model is shown in Fig. \ref{fig:model}. Further details of the architecture modifications are outlined in Section \ref{sec:model}.

\subsection{Machine annotation} \label{sec:unlabelled}
The training of trajectory forecasting models in a supervised learning setting requires dense (per-frame) bounding box annotation of pedestrians, which are expensive to obtain by hand.  For this reason, the number and size of datasets with densely annotated pedestrian bounding boxes is limited. The size of existing datasets \cite{context-based,jaad} is prohibitive for the training of high-capacity deep learning models, which rely on large quantities of data to learn an effective feature representation. To overcome this issue, we propose to learn from unlabeled data by using an automated pedestrian detection and tracking algorithm to generate bounding boxes without human labor.

Given an input video sequence, pedestrian detection algorithms obtain an estimate of the location $L_t$ for each pedestrian, and a tracking method then links these estimated locations across each timestep $t$. Given a set of such detections, we adopt the self-supervision learning paradigm by training our model to predict future pedestrian locations, $L_{t+n}$, given only the current and past locations, viz., $L_{t-m} \ldots L_t$.

A similar annotation process is proposed in \cite{fpl}, in which pedestrians are detected and tracked using \cite{openpose}. However, automated detectors do not perform on par with human annotators, and make different errors to humans, such as false positive detections of vertical structures \cite{howfarfromsolving}. Due to this, it is not evident that models trained on machine-annotated data will generalize across datasets and to human-annotated data. To verify our proposed machine-annotation regime, we validate the performance of our model on a human-annotated dataset. We adopt the conventional methodology of pre-training on a large dataset before fine-tuning on a smaller target dataset, intending to improve generalizability on the target dataset \cite{howtransferable}.

\section{EXPERIMENTS} \label{sec:4}

\subsection{Datasets}
\begin{figure}[t]
   \centering
   \includegraphics[width=.90\linewidth]{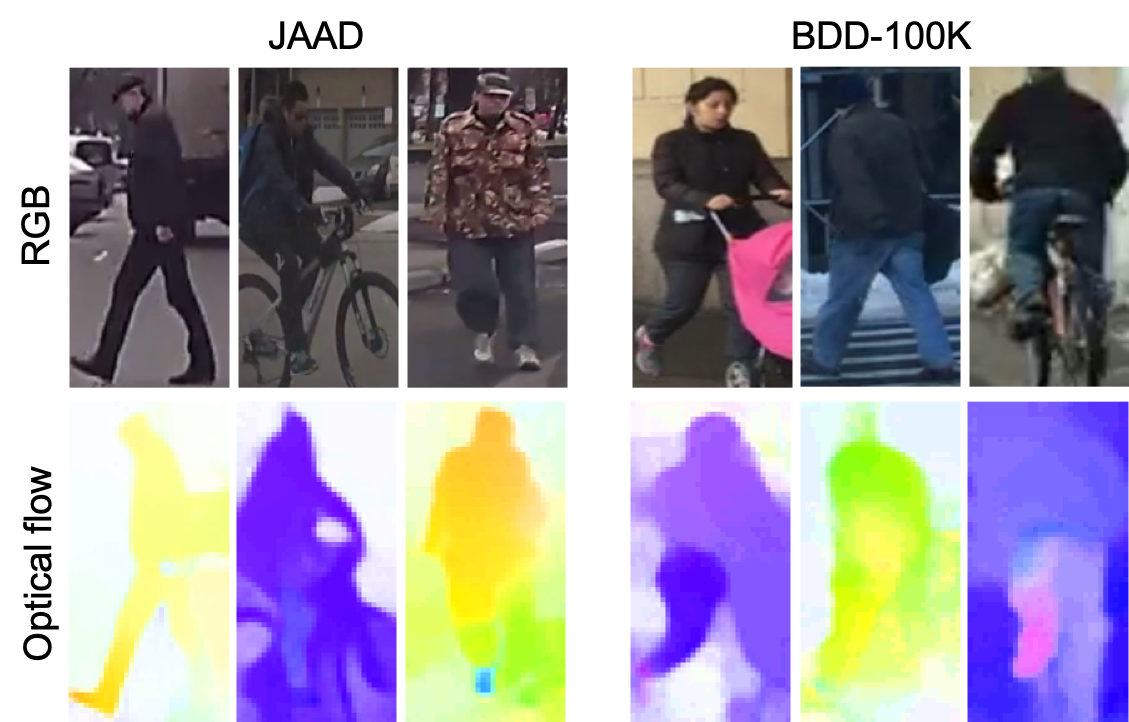}
     \caption{Example pedestrians with associated optical flow obtained using Flownet2-CSS. Left 3 images are human-annotated pedestrians from the JAAD dataset, right 3 images are pedestrians detected on the BDD-100k dataset using YOLOv3. Optical flow captures motion resulting from both the camera and pedestrian.}
     \label{fig:flow}
       \vspace{-4mm}
\end{figure}

We use two datasets in our experiments, JAAD \cite{jaad} and BDD-100K \cite{bdd100k}. Both datasets consist of videos captured by a front-facing camera mounted behind a windshield collected by cars driving on public roads in Europe and North America. The JAAD dataset contains dense pedestrian bounding box annotation, that is, annotations are provided for each frame. BDD-100K, however, contains sparse bounding box annotation. Only one frame per video is annotated. We do not use the sparely annotated bounding boxes. Due to the huge number of videos in BDD-100K, we use only the first 10,000 videos. This subset is henceforth referred to as BDD-10K.  Example pedestrian images from both datasets are shown in Fig. \ref{fig:flow} (first row). Videos from the JAAD dataset are downsampled with bilinear interpolation to match the BDD-10K dataset resolution of $1280 \times 720$. The frame rate of both datasets is downsampled from 30 to 15 frames per second to reduce redundancy between consecutive frames.

% \begin{table}[h]
% \caption{An Example of a Table}
% \label{table_example}
% \begin{center}
% \begin{tabular}{|c||c|}
% \hline
% One & Two\\
% \hline
% Three & Four\\
% \hline
% \end{tabular}
% \end{center}
% \end{table}

\begin{figure*}[t]
\begin{center}
\includegraphics[height=9.9cm,width=0.85\linewidth]{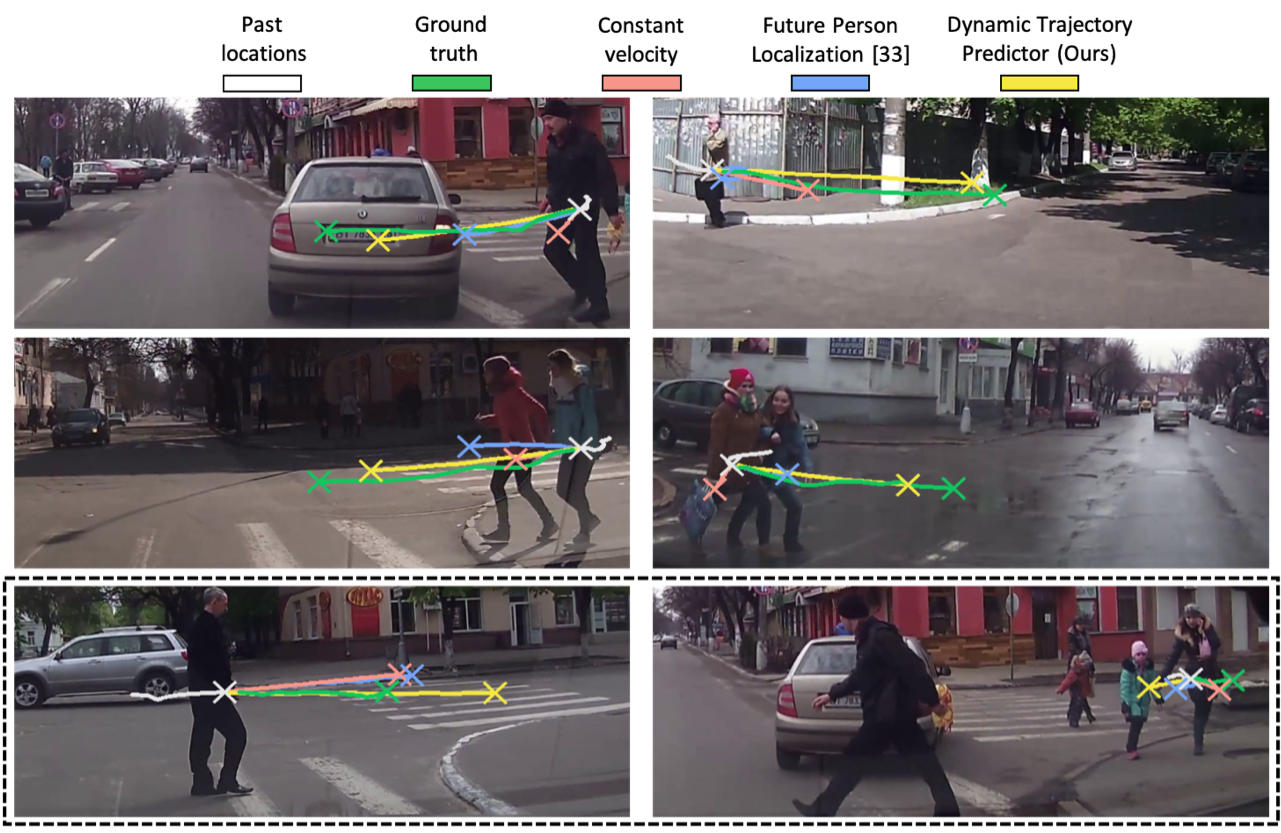}
\end{center}
  \vspace{-4mm}
   \caption{Example successful (top 2 rows) and unsuccessful (bottom row) trajectory forecasts on the JAAD test set. See main text for discussion. Best viewed in color.}
\label{fig:qualitative}
  \vspace{-3mm}
\end{figure*}

\subsection{Dynamic Trajectory Predictor} \label{sec:model}

\textbf{Implementation.} To evaluate DTP, we use the JAAD dataset. Pedestrians smaller than 50 pixels in height, occluded pedestrians, and tracks shorter than 25 frames are discarded. Optical flow is extracted from cropped pedestrians using the provided human-annotated bounding boxes with the Flownet2-CSS algorithm \cite{FlowNet2}. Pixel displacements are clipped at $\pm 50$ and scaled to the range $[0,1]$. Example pedestrian flow images are shown in Fig. \ref{fig:flow} (second row).

We use a stack of $m$ horizontal and $m$ vertical optical flow frames at timesteps $t-m$ to $t$. Features are computed from the $2m$ input channels using the ResNet-18 CNN architecture \cite{resnet}. We modify the first convolutional layer to use $2m$ input channels rather than 3, keeping other dimensions the same. We replace the 1000-D softmax output layer with a 30-D fully connected later which produces predictions for the $x$ and $y$ coordinates of the 15 future bounding box centroids.  We use cross-modality pre-training and partial batch normalization \cite{tsn} to initialize our CNN with ImageNet weights. The model is optimized to minimize the $\mathcal{L}_2$ loss between the true and predicted future locations, $L_{t+1} \ldots L_{t+n}$ and $\hat{L}_{t+1} \ldots \hat{L}_{t+n}$, and is trained until convergence using the Adam \cite{adam} optimizer with an initial learning rate of $10^{-5}$, which is reduced to $10^{-6}$ once performance saturates. We use a batch size of 64 and a weight decay of $10^{-2}$. Each pedestrian is resized to $256 \times 256$ pixels. For data augmentation, a randomly cropped sub-image of size $224 \times 224$ is taken.

We split the JAAD dataset into training (videos 0-250) and testing (videos 251-346) sets. We perform 5 fold cross-validation on the JAAD training set to tune hyperparameters. Once hyperparameters are fixed, we obtain an estimate of the model's generalizability by training on each of the 5 folds until performance on the respective validation set saturates. We then evaluate the model on the test set. We report the mean performance on the test set with associated 95\% confidence intervals for the 5 folds.

\textbf{Evaluation.} We use two metrics to evaluate model performance, mean squared error (MSE) and displacement error (DE@$t$) at timesteps up to 15, following \cite{longterm,fpl}. The MSE is the mean of the squared errors of the predicted centroid in pixels from all timesteps 1 to $n$ and across all samples in the test set. The DE@$t$ is the mean Euclidean distance in pixels of the predicted and ground truth centroid for timestep $t$ only. Both metrics are relative to an image resolution of $1280 \times 720$.

We evaluate our proposed approach with 4 different inputs: a single RGB frame at time $t$, a single optical flow frame at time $t$, a stack of 5 optical flow frames at times $t-4$ to $t$, and a stack of 9 optical flow frames at times $t-8$ to $t$. We use 9 as our maximum value of $m$ rather the then 10 frames commonly used for action recognition \cite{twostream,tsn} for a fair comparison with Future Person Localization (FPL) \cite{fpl}, which uses 10 frames as input. As each optical flow frame requires two consecutive RGB frames to be computed, using 10 input frames results in 9 optical flow frames. Following prior works \cite{fpl,sociallstm} we adopt constant velocity (CV) and constant acceleration (CA) as baselines. For the CV baseline, we compute the average velocity in the image space using the previous locations and predict the future location assuming the pedestrian maintains a linear velocity. Similarly, for the CA baseline, we compute the average acceleration using the previous locations and extrapolate these values into the future timesteps assuming linear acceleration. Using 4 previous locations resulted in the best cross-validation performance.

\textbf{Results.}  Table \ref{tab:inputs} shows the performance of each model with different input modalities in comparison with the CV and CA baselines. Due to the relatively poor performance of the RGB input, we do not fuse RGB and optical flow models as in the two-stream model \cite{twostream}. Example outputs of our model using a stack of 9 optical flow frames compared to baselines are shown in Fig. \ref{fig:qualitative}. DTP performs particularly well in situations where a pedestrian first begins walking, and when the ego-vehicle begins to turn sharply (top two rows). DTP performs less well under conditions of significant background motion such as those due to other vehicles (bottom left image) or upper body motion in the counter walking direction  (bottom right image).

We compare our method using a stack of 9 optical flow frames with linear baselines and FPL \cite{fpl} in Table \ref{tab:model}. We modify FPL to output 15 timesteps into the future rather than the 10 as in the original architecture and use optical flow for ego-motion estimation as described in \cite{fpl}. Both DTP and FPL see a reduction in error with our proposed CV correction term $C_t$ (rather than directly predicting the location displacement $\Delta L_{t+n}$). DTP attains the best performance.

\begin{table}[ht]
\begin{center}   \caption{Input modality comparision.}
  \begin{tabular}{ c  c  c  c  c}
    \hline
    \textbf{Input modality} & \textbf{MSE} & \textbf{DE@5}  &  \textbf{DE@10}  &  \textbf{DE@15}  \\
    \hline
    CA & $1426$ & 15.3 & 28.3 & $52.8$ \\
    CV & $1148$ & 16.0 & 26.4 & $47.5$ \\ \hline
    RGB frame  & $1042$ & 11.6 & 24.9 & $45.2$\\
    Optical flow frame  & $873$ & 11.1 & 23.0 & $41.2$\\
    5 optical flow frames  & $651$ & 9.4 & 19.3 & $35.6$\\
    \textbf{9 optical flow frames}  & $\mathbf{610}$ & $\mathbf{9.2}$ & $\mathbf{18.7}$ & $\mathbf{34.6}$\\
    \hline
  \end{tabular}\label{tab:inputs}
  \end{center}
    \vspace{-6mm}
  \end{table}

\begin{table}[ht]
\begin{center}   \caption{Model comparision.}
  \begin{tabular}{ c  c  c  c }
    \hline
    \textbf{Model} & \makecell{\textbf{CV correction} \\ \textbf{term}} & \textbf{MSE} & \textbf{DE@15}  \\
    \hline
    FPL \cite{fpl} & \xmark & $1405 \pm 182$ & $49.5 \pm 2.9 $\\
    FPL \cite{fpl} & \cmark & $881 \pm 44$ & $41.3 \pm 1.2$\\
    DTP & \xmark &$1404 \pm 94$\hphantom{2} & $54.6 \pm 2.6 $\\
    \textbf{DTP}  & \cmark  & $\mathbf{610 \pm 21}$ & $\mathbf{34.6 \pm 0.5}$\\
    \hline
  \end{tabular}\label{tab:model}
  \end{center}
  \vspace{-5mm}
  \end{table}

\subsection{Machine annotation}

\textbf{Implementation.} We annotate pedestrian bounding boxes in the BDD-10K dataset using two popular off-the-shelf object detectors, YOLOv3 \cite{yolo3} and Faster-RCNN \cite{faster}. Although the detectors are capable of detecting a variety of objects, we use the pedestrian class only. Our aim here is to evaluate the robustness of our proposed system to multiple automated detectors, rather than to compare detector performance directly. Nonetheless, for consistency, we train both detectors on the same dataset (MS-COCO \cite{coco}) and threshold confidence scores at $0.6$.

Once frame-wise detections are obtained, detections are associated across frames using the Deepsort \cite{deepsort} tracking-by-detection algorithm resulting in a series of bounding boxes and tracking identifiers. We use the same setup as the JAAD dataset and discard detections with height fewer than 50 pixels, and tracks shorter than 25 frames. Using this annotation scheme, we find a total of 16,900 valid non-overlapping pedestrian tracks using YOLOv3 and 13,200 using Faster-RCNN.

\textbf{Evaluation.} We use an 80\%-20\% training-validation split for BDD-10K. We pre-train DTP on BDD-10K using the same hyperparameters as outlined in Section \ref{sec:model}. Once performance on the validation set saturates, the model is fine-tuned on the JAAD training set. We evaluate the trajectory forecasting performance with and without pre-training rather than the pedestrian detection quality, owing to the lack of human-annotated bounding boxes.

\textbf{Results}. The impact of machine-annotated pre-training using the YOLOv3 detector before fine-tuning on the human-annotated JAAD dataset is shown in Table \ref{tab:pre-training}.

\vspace{-2mm}
\begin{table}[ht]
\begin{center}   \caption{Impact of pre-training on BDD-10K with YOLOv3.}
  \begin{tabular}{ c  c c  c }
    \hline
    \textbf{Model} &\makecell{\textbf{Pre-training with} \\ \textbf{machine annotation}} & \textbf{MSE} & \textbf{DE@15}  \\
    \hline
    FPL \cite{fpl} & \xmark & $881 \pm 44$ & $41.3 \pm 1.2 $\\
    FPL \cite{fpl} & \cmark & $805 \pm 46$ & $40.1 \pm 1.2$ \\
    DTP & \xmark  & $610 \pm 21$ & $34.6 \pm 0.5$\\
    \textbf{DTP}  & \cmark  & $\mathbf{539 \pm 13}$ & $\mathbf{32.7 \pm 0.4}$\\
    \hline
  \end{tabular}\label{tab:pre-training}
    \vspace{-2mm}

  \end{center}
  \end{table}

\begin{figure}[t]
\begin{center}
\includegraphics[width=0.95\linewidth,height=5.6cm]{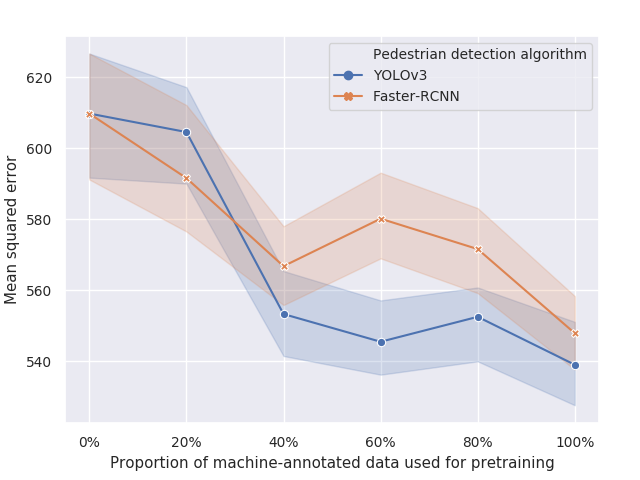}
\end{center}
    \vspace{-3mm}

   \caption{Impact of pre-training dataset size and pedestrian detection algorithm on the performance on JAAD test set. Shaded areas show the 95\% confidence interval.}
\label{fig:pre-training}
    \vspace{-5mm}
\end{figure}

We evaluate the impact of pre-training dataset size and pedestrian detector by training on subsets of BDD-10K ranging from 20\% to 100\% of the total dataset size. Fig. \ref{fig:pre-training} shows the MSE on the JAAD test set for both YOLOv3 and Faster-RCNN. In general, the error on the JAAD test set reduces as larger subsets of our machine-annotated dataset, BDD-10k, is used for pre-training. The reduction in error may be due to the model's ability to learn the motion patterns of under-represented classes, such as children or the elderly, from a larger dataset.

\section{CONCLUSION}

We have presented a model and complementary machine annotation scheme for pedestrian trajectory forecasting from onboard a moving vehicle. Our model, DTP, forecasts trajectory for time horizons up to one second by anticipating a change in velocity using optical flow information. By introducing a method for annotating data without human labor, DTP and other similar models may leverage large-scale datasets for learning effective feature representations.

\section*{ACKNOWLEDGMENT}

%%%%%%%%%%%%%%%%%%%%%%%%%%%%%%%%%%%%%%%%%%%%%%%%%%%%%%%%%%%%%%%%%%%%%%%%%%%%%%%%

This work is funded by the UK EPSRC (grant no. EP/L016400/1) and the EU Horizon 2020 project IDENTITY (Project No. 690907). Portions of this work were done when Styles was at MSU. Our thanks to NVIDIA for supporting this research with their generous hardware donation.

{\small
\bibliographystyle{ieee}
\bibliography{ivbib}
}

\end{document}